\documentclass{article}
\usepackage{enumitem}

\usepackage[final]{neurips_2026}

\usepackage[utf8]{inputenc}
\usepackage[T1]{fontenc}
\usepackage{hyperref}
\usepackage{url}
\usepackage{booktabs}
\usepackage{amsfonts}
\usepackage{amsmath}
\usepackage{nicefrac}
\usepackage{microtype}
\usepackage[table]{xcolor}
\usepackage{graphicx}
\usepackage{multirow}
\usepackage{subcaption}

\newcommand{\ours}{MARS}
\newcommand{\mask}{\texttt{[MASK]}}

\title{MARS: Enabling Autoregressive Models Multi-Token Generation}

\author{%
  Ziqi Jin$^{1}$ \quad Lei Wang$^{2}$ 
  \quad Ziwei Luo$^3$ \quad Aixin Sun$^1$ \\[1mm]  
  \\
  $^1$Nanyang Technological University \quad $^2$Singapore Management University \quad $^3$Uppsala University \\[1mm]
}
\begin{document}

\maketitle

\begin{abstract}
Autoregressive (AR) language models generate text one token at a time,
even when consecutive tokens are highly predictable given earlier context.
We introduce \textbf{\ours{}} (\textbf{M}ask \textbf{A}uto\textbf{R}egre\textbf{S}sion),
a lightweight fine-tuning method that teaches an instruction-tuned AR model
to predict multiple tokens per forward pass.
\ours{} adds no architectural modifications, no extra parameters, and
produces a single model that can still be called exactly like the original
AR model with no performance degradation.
Unlike speculative decoding, which maintains a separate draft model alongside
the target, or multi-head approaches such as Medusa, which attach additional
prediction heads, \ours{} requires only continued training on existing
instruction data.
When generating one token per forward pass, \ours{} matches or exceeds
the AR baseline on six standard benchmarks.
When allowed to accept multiple tokens per step, it maintains
baseline-level accuracy while achieving 1.5--1.7$\times$ throughput.
We further develop a block-level KV caching strategy for batch inference,
achieving up to 1.71$\times$ wall-clock speedup over AR with KV cache
on Qwen2.5-7B.
Finally, \ours{} supports real-time speed adjustment via confidence
thresholding: under high request load, the serving system can
increase throughput on the fly without swapping models or restarting,
providing a practical latency--quality knob for deployment. \footnote{Our code is available at: \url{https://github.com/Xalp/MARS}}
\end{abstract}

\section{Introduction}
\label{sec:introduction}

Autoregressive (AR) language models spend the same compute on every token:
one forward pass produces exactly one token, whether that token is the
inevitable ``\texttt{the answer is}'' following a multiple-choice prompt or
a genuinely uncertain next word in an open-ended generation.
This uniform cost is wasteful.

Existing approaches to multi-token generation all modify the deployment
stack. Speculative decoding~\citep{chen2023accelerating, leviathan2023fast}
maintains a separate draft model alongside the target, doubling memory
footprint and adding orchestration complexity. Multi-head approaches such as
Medusa~\citep{cai2024medusa} and EAGLE~\citep{li2024eagle} attach additional
prediction heads to the architecture, requiring extra parameters and
head-specific training. Both families introduce components beyond the
original model, complicating production serving pipelines.

In this work we take a different angle: rather than building auxiliary
components around an AR model, we ask whether lightweight fine-tuning alone
can give the model \textit{an additional capability to generate multiple tokens per
forward pass}, while preserving its original behavior as a standard AR model.
The resulting model should be a strict superset: it can still be served
exactly like the original, producing one token at a time with no quality
loss, but it also optionally accepts multiple tokens per step when confident,
accelerating generation with minimal cost.

A natural starting point is block masked diffusion, which trains the model to
predict multiple tokens per step within a fixed-size block. However, existing
AR-to-block-diffusion conversions result in substantial quality degradation,
particularly on reasoning and coding tasks ~\citep{arriolablock, zhou2026dllmsimplediffusionlanguage, gongscaling, fu2025efficient}.
In Section~\ref{sec:gap_analysis}, we identify four gaps between AR and
block diffusion that may account for this degradation, but find that only
one is inherent to multi-token prediction: the use of mask tokens as
placeholders for unknown future tokens. The other three arise from design
choices that unnecessarily depart from the original AR model, and can be
fully closed.

From these motivations, we introduce \textbf{\ours{}} (\textbf{M}ask \textbf{A}uto\textbf{R}egre\textbf{S}sion). \ours{} maintains causal intra-block
attention, uses right-shifted logits, and generates strictly left-to-right,
leaving only the inherent masking gap. As shown in Figure~\ref{fig:case_study},
the resulting model adaptively generates multiple tokens per forward pass when
confident, while falling back to single-token generation for novel content.
When generating one token per step, \ours{} matches or exceeds the original
AR baseline on six benchmarks, meaning there is \textit{no quality cost} for
gaining the option to accelerate. When multi-token generation is enabled,
\ours{} achieves 1.5--1.7$\times$ throughput while maintaining
baseline-level accuracy.

\begin{figure}
\centering
\small
\begin{tabular}{p{0.95\linewidth}}
\toprule
\textbf{Question:} Toulouse has twice as many sheep as Charleston. Charleston has 4 times as many sheep as Seattle. How many sheep do Toulouse, Charleston, and Seattle have together if Seattle has 20 sheep? \\
\midrule
\textbf{\ours{}-7B (confidence threshold $\tau{=}0.95$), 66 forwards, 2.55 tokens/forward:} \\[3pt]
Let 's \colorbox{blue!16}{break} \colorbox{blue!16}{it} \colorbox{blue!16}{down} \colorbox{blue!16}{step} \colorbox{blue!16}{by} \colorbox{blue!16}{step:}
1 . \colorbox{blue!16}{Seattle has} \colorbox{blue!24}{20 sheep.}
\colorbox{blue!24}{2. Charleston has} \colorbox{blue!24}{4 times as} \colorbox{blue!24}{many sheep as Seattle} \colorbox{blue!24}{, so Charleston has}
4 \colorbox{blue!16}{x 20} \colorbox{blue!16}{= 80} \colorbox{blue!16}{sheep.}
\colorbox{blue!16}{3.} \colorbox{blue!24}{Toulouse has twice} \colorbox{blue!24}{as many sheep as} \colorbox{blue!24}{Charleston, so T} \colorbox{blue!16}{oulouse has 2} \colorbox{blue!16}{x 80} \colorbox{blue!16}{= 16} 0 sheep.
\colorbox{blue!16}{Now} \colorbox{blue!16}{, let} \colorbox{blue!16}{'s add} \colorbox{blue!16}{up} \dots{}
\colorbox{blue!24}{Total number of} \colorbox{blue!24}{sheep} \colorbox{blue!16}{: 20} \colorbox{blue!16}{+ 80} \colorbox{blue!16}{+ 16} \colorbox{blue!16}{0 = 1} \colorbox{blue!16}{00 +} \colorbox{blue!16}{160 =} \colorbox{blue!16}{260} \colorbox{blue!16}{sheep} \\[3pt]
{\footnotesize \textcolor{gray}{Shaded regions = multiple tokens accepted in one forward pass. Unshaded tokens are generated one at a time. Darker = more tokens per step. Formulaic phrases and calculations are chunked; novel reasoning proceeds token-by-token.}} \\
\bottomrule
\end{tabular}
\caption{Example \ours{} generation on GSM8K. The model adaptively generates 1--4 tokens per forward pass based on confidence: predictable continuations are batched together, while novel content proceeds token-by-token. This achieves 2.55$\times$ token per forward over standard AR decoding.}
\label{fig:case_study}
\end{figure}

\paragraph{Contributions.}
\begin{itemize}
    \item We analyze the four gaps between AR and block-masked prediction and show that three of them are eliminable design choices rather than inherent limitations. Closing these gaps is sufficient to recover baseline quality without architectural changes.

    \item We propose \ours{}, a lightweight fine-tuning method that teaches
    an instruction-tuned AR model to optionally predict multiple tokens per
    forward pass, with no architectural changes, no additional parameters,
    and reusing the same SFT data. The resulting model can be used exactly
    like the original AR model with no quality loss, or switched to
    multi-token mode for faster generation on the fly.

    \item We identify an auxiliary SFT loss on the clean input stream as
    the key ingredient for preserving performance at larger block sizes,
    where the fraction of AR-like training signal would otherwise decay.

    \item We develop a block-level KV caching strategy for batch inference and
    demonstrate wall-clock speedups of up to 1.71$\times$ over AR with
    KV cache on Qwen2.5-7B.
\end{itemize}

\section{Background and Related Work}
\label{sec:preliminary}

An autoregressive (AR) language model generates text by predicting one token at a time, each conditioned on all previous tokens via causal (left-to-right) attention. Generating $T$ tokens requires $T$ serial forward passes, so inference cost scales linearly with output length regardless of how predictable individual tokens are. If we want a single forward pass to predict multiple future tokens using the same backbone and language model head, a natural construction is \textit{block-masked prediction}: replace a contiguous block of $B$ future tokens with \mask{} placeholders and train the model to recover them, conditioned on the clean tokens from all preceding blocks. To train all blocks in parallel within a single forward pass, we concatenate the clean and masked sequences: $[\mathbf{x}; \tilde{\mathbf{x}}]$, with a structured attention mask that controls visibility (detailed in Section~\ref{sec:method_training}). However, directly applying this to a pretrained AR model introduces a mismatch: the model was trained under purely causal, fully visible context, but now must predict from partially masked input with potentially different attention patterns and generation order. This mismatch is the source of the quality degradation observed in prior work~\citep{arriolablock}, and motivates the analysis in Section~\ref{sec:gap_analysis}.

\paragraph{Discrete diffusion and masked language models.}
Diffusion models for discrete data~\citep{austin2021structured, li2022diffusion} generate tokens through iterative denoising. Early non-autoregressive work in machine translation~\citep{gu2018non, ghazvininejad2019mask} demonstrated significant speedups at the cost of quality. More recently, masked diffusion language models~\citep{sahoo2024simple, shi2024simplified, lou2024discrete, gat2024discrete, ye2025dream, nielarge} apply iterative unmasking to text generation and have begun to close the quality gap with AR models. These models typically use bidirectional attention, a design choice inherited from diffusion models in vision. Several works have converted pretrained AR models into diffusion language models~\citep{arriolablock, zhou2026dllmsimplediffusionlanguage, gongscaling, fu2025efficient}, demonstrating that the transition is feasible without training from scratch, though often with significant quality degradation due to the mismatch between the AR model's learned causal attention pattern and the bidirectional attention adopted during conversion.

\paragraph{Causality and multi-token generation.}
A growing body of work shows that causal attention does not conflict with multi-token parallel generation. Block Diffusion~\citep{arriolablock} uses causal attention across blocks while keeping bidirectional attention within each block. Building on this, several works further accelerate diffusion LM inference~\citep{wu2025fast}, including Fast-dLLM which introduces KV caching for bidirectional diffusion models, and self-speculative decoding~\citep{gao2025self} which uses the diffusion model itself as both drafter and verifier. CARD~\citep{ruan2026causal} applies strictly causal attention to the entire sequence and finds that this actually \textit{improves} over bidirectional alternatives. ARMD~\citep{karami2026auto} and A3~\citep{du2026autoregressive} similarly adopt causal structure for masked diffusion and any-order generation, respectively. Eso-LM~\citep{sahoo2026esotericlanguagemodelsbridging} further unifies AR and masked diffusion under causal attention with KV caching. Together, these results establish that bidirectional attention in diffusion language models was a design choice inherited from vision, not a requirement for parallel token generation.

This observation motivates a natural question: if causality and multi-token generation are compatible, how much modification does an AR model actually need to generate multiple tokens at once? Our gap analysis (Section \ref{sec:gap_analysis}) suggests the answer is surprisingly little. Of the four gaps between AR and block-masked prediction, three turn out to be eliminable design choices. The sole inherent gap is some positions must predict from incomplete context, which can be addressed with a simple fine-tuning objective.

\paragraph{Multi-token prediction and parallel decoding.}
Multi-token prediction (MTP)~\citep{gloeckle2024better} trains AR models with $k{-}1$ auxiliary heads to predict future tokens in parallel; DeepSeek-V3~\citep{liu2024deepseek} deploys this at 671B scale. Medusa~\citep{cai2024medusa} and EAGLE~\citep{li2024eagle} add lightweight heads for speculative multi-token proposals. These methods require additional parameters and architectural modifications. \ours{} achieves multi-token prediction without extra heads, using the same language model head for all positions within a block. Speculative decoding~\citep{leviathan2023fast, chen2023accelerating} avoids modifying the target model but requires maintaining a separate draft model.

Jacobi decoding~\citep{tengaccelerating} and Lookahead decoding~\citep{fu2024break} achieve parallel generation from a single unmodified model by treating AR generation as a fixed-point iteration. CLLMs~\citep{kou2024cllms} improve upon this by training the model to converge faster under Jacobi iteration. These methods initialize future positions with \textit{random tokens} and rely on convergence from incorrect prefixes. \ours{} takes a complementary approach: instead of random initialization, we use \mask{} tokens as explicit placeholders and train the model to predict from this incomplete context directly. This yields substantially higher acceptance rates (1.5$\times$ vs 1.07$\times$ tokens per forward for Jacobi; Appendix~\ref{app:jacobi}).

\section{Method}
\label{sec:method}

The design of \ours{} is guided by one principle: {\textbf{the model must remain a fully functional AR model}}. Multi-token prediction is an added capability, not a replacement. We begin by analyzing where prior block-masked approaches break this property, then show how \ours{} closes every eliminable gap at training and inference.

\subsection{Where Does Block-Masked Prediction Fail?}
\label{sec:gap_analysis}

Block-masked prediction fails when it departs from the autoregressive model along multiple axes. We identify four gaps between AR and block diffusion (Table~\ref{tab:gap}). Of these, one is inherent to multi-token prediction and cannot be avoided; the other three are eliminable design choices that unnecessarily break AR compatibility.

\begin{table}[t]
\centering
\caption{Gaps between AR and block diffusion. \ours{} aligns with AR on gaps (2)--(4), leaving only the inherent masking gap.}
\label{tab:gap}
\small
\newcommand{\cmatch}[1]{\cellcolor{green!12}#1}
\newcommand{\cmiss}[1]{\cellcolor{red!12}#1}
\begin{tabular}{lccc}
\toprule
 & AR & \ours{} & Block Diffusion \\
\midrule
(1) Token Masking & \cmatch{No masking} & \cmiss{Masked within block} & \cmiss{Masked within block} \\
(2) Attention Pattern & \cmatch{Causal} & \cmatch{Causal} & \cmiss{Bidirectional (intra-block)} \\
(3) Logits Alignment & \cmatch{Right-shifted} & \cmatch{Right-shifted} & \cmiss{Varies} \\
(4) Generation Order & \cmatch{Left-to-right} & \cmatch{Left-to-right} & \cmiss{Confidence-based} \\
\bottomrule
\end{tabular}
\end{table}

Gap (1), \textit{token masking}, is inherent: to predict multiple tokens in parallel, future positions must be replaced with \mask{} placeholders. This is the fundamental cost of multi-token prediction. The remaining three gaps are eliminable design choices in prior systems that unnecessarily break AR compatibility. Gap (2), \textit{attention pattern}: some block diffusion approaches~\citep{arriolablock} use bidirectional attention within blocks, but a bidirectional model is no longer a functional AR model; \ours{} keeps strictly causal attention everywhere. Gap (3), \textit{logits alignment}: AR models predict $x_{t+1}$ from position $t$ (right-shifted logits); changing this convention breaks the output head's AR function, so \ours{} preserves it. Gap (4), \textit{generation order}: confidence-based diffusion methods unmask tokens out of order within a block, breaking left-to-right generation; \ours{} always accepts tokens strictly left-to-right. By closing gaps (2)--(4), the resulting model remains a fully functional AR model, with the only difference being that it sees \mask{} placeholders within the current block.

\subsection{Training}
\label{sec:method_training}

\ours{} starts from an AR SFT checkpoint trained on the target instruction-following data with standard next-token prediction. Starting from this checkpoint ensures the model has already absorbed the training data distribution, so that \ours{} training focuses solely on learning the masked prediction paradigm (see Section~\ref{sec:exp_setup} for details).

The high-level idea is simple: we run two copies of the sequence through the model in parallel. The \textit{clean stream} keeps the original tokens intact and trains the model with ordinary AR next-token prediction. The \textit{noisy stream} replaces each block of $B$ tokens with \mask{} placeholders and asks the model to predict them, using the clean prefix from earlier blocks as context. Both streams share the same forward pass through a structured attention mask that enforces the correct visibility for each position. This design closes gaps (2) and (3) from Section~\ref{sec:gap_analysis} by construction: attention is strictly causal everywhere, and logits are right-shifted to match the AR convention.

Given a response sequence $\mathbf{x} = (x_1, \ldots, x_L)$, we divide it into blocks of size $B$ and replace all tokens in each block with \mask{}:
\begin{equation}
    \tilde{\mathbf{x}} = (\underbrace{\mask{}, \ldots, \mask{}}_{B}, \underbrace{\mask{}, \ldots, \mask{}}_{B}, \ldots)
\end{equation}
The model processes a concatenated input $\mathbf{z} = [\mathbf{x}; \tilde{\mathbf{x}}]$ of length $2L$, where the first $L$ positions are the clean stream and the last $L$ are the noisy stream.

We define the attention mask $\mathbf{M} \in \{0, -\infty\}^{2L \times 2L}$ over $\mathbf{z}$. Let $\beta(t) = \lceil t / B \rceil$ denote the block index of position $t$ within either stream. For query position $i$ and key position $j$:
\begin{equation}
    M_{ij} = \begin{cases}
    0 & \text{if } i, j \leq L,\; j \leq i \quad \text{(clean causal)} \\
    0 & \text{if } i, j > L,\; \beta(i{-}L) = \beta(j{-}L),\; j \leq i \quad \text{(noisy intra-block causal)} \\
    0 & \text{if } i > L,\; j \leq L,\; \beta(j) < \beta(i{-}L) \quad \text{(noisy} \to \text{clean cross-stream)} \\
    -\infty & \text{otherwise}
    \end{cases}
    \label{eq:attention_mask}
\end{equation}
The \textit{clean causal} case gives the clean stream (positions $1, \ldots, L$) standard causal self-attention, identical to AR training. The \textit{noisy intra-block causal} case allows each noisy position to attend causally within its own block (only \mask{} tokens, so only positional information flows). The \textit{cross-stream} case allows each noisy block $k$ to see clean tokens from blocks $1, \ldots, k{-}1$, providing the prefix context needed for prediction.

\begin{figure}[t]
    \centering
    \includegraphics[width=\textwidth]{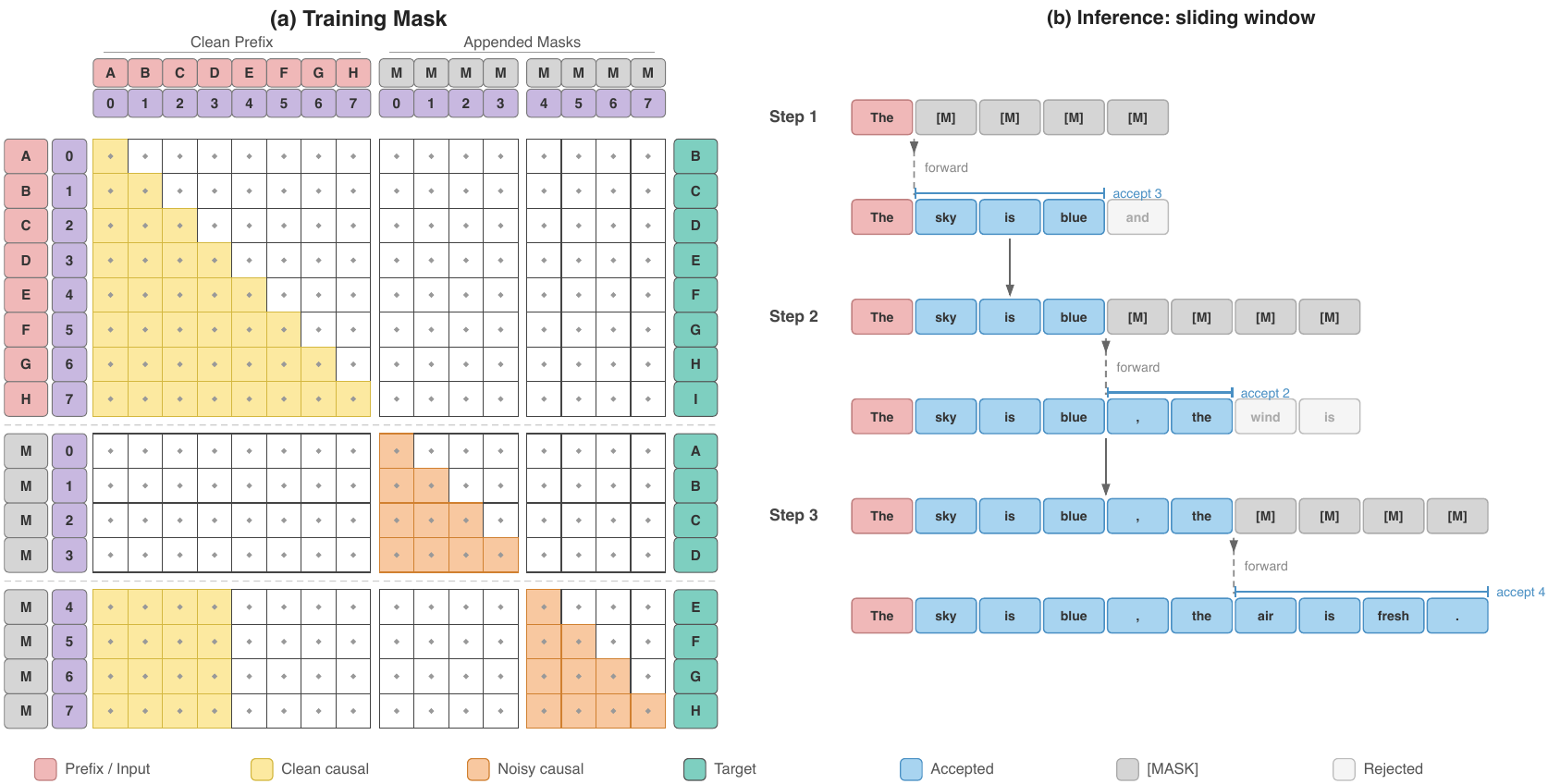}
    \caption{\ours{} attention mask and inference for $L{=}8$, $B{=}4$. \textbf{Left:} training mask with $[\mathbf{x} \mid \tilde{\mathbf{x}}]$ concatenation. The orange cells show that noisy positions attend to each other \textit{causally} within each block, in contrast to Block Diffusion~\citep{arriolablock} which uses bidirectional attention within blocks. \textbf{Right:} sliding-window inference. The dashed line marks the generation cursor; $B$ \mask{} tokens are appended and filled via one forward pass. Accepted tokens (blue) slide into the prefix for the next step.}
    \label{fig:attention_mask}
\end{figure}

The training loss on the noisy stream is cross-entropy over masked positions:
\begin{equation}
    \mathcal{L}_{\text{mask}} = -\sum_{t \in \mathcal{M}} \log p_\theta(x_t \mid \tilde{\mathbf{x}}, \mathbf{x}_{<\beta(t)})
    \label{eq:mask_loss}
\end{equation}
where $\mathcal{M}$ denotes the set of masked positions and $\mathbf{x}_{<\beta(t)}$ denotes clean tokens from all blocks preceding block $\beta(t)$.

\subsection{Preserving Autoregressive Competence}
\label{sec:signal_decay}

Closing gaps (2)--(4) ensures that \ours{} \textit{behaves} like an AR model at inference. But training must also ensure the model \textit{remains} an AR model in terms of capability. Block-masked training, by itself, gradually erodes the AR signal, and this erosion gets worse exactly when larger blocks would be most useful. The SFT loss is not a regularization trick; it is the mechanism that keeps the model's AR competence intact while it learns block prediction.

Within a block of size $B$, position $t$ (1-indexed within the block) is conditioned on $t{-}1$ masked tokens from the same block, plus the fully clean prefix from prior blocks. Only the first position ($t{=}1$) sees entirely clean context, making its prediction exactly equivalent to AR next-token prediction. Positions $t{=}2, \ldots, B$ see progressively more \mask{} tokens in place of real context, with position $B$ seeing $B{-}1$ placeholders. As a simple proxy for how much AR-like signal the model receives, we count the fraction of positions with \textit{fully clean} context:
\begin{equation}
    r_{\text{AR}}^{\text{(mask only)}} = \frac{1}{B}
    \label{eq:signal_ratio_no_sft}
\end{equation}
This is a coarse measure---positions near the start of a block still receive mostly clean context---but it captures the trend: as $B$ grows, the training signal becomes increasingly unlike standard AR. For $B{=}4$ the ratio is 25\%; for $B{=}8$, 12.5\%; for $B{=}16$, just 6.25\%. Empirically, this decay leads to significant degradation on reasoning and coding tasks at larger block sizes (Table~\ref{tab:main_results}).

The clean stream in $\mathbf{z} = [\mathbf{x}; \tilde{\mathbf{x}}]$ already uses standard causal attention on the original tokens, and its logits are computed during the forward pass at no extra cost. By adding an AR loss on these logits, we ensure the model never stops practicing next-token prediction, no matter how large the block size:
\begin{equation}
    \mathcal{L} = \mathcal{L}_{\text{mask}} + \mathcal{L}_{\text{AR}}
    \label{eq:total_loss}
\end{equation}
where $\mathcal{L}_{\text{AR}} = -\sum_{t=1}^{L} \log p_\theta(x_t \mid \mathbf{x}_{<t})$ is the standard next-token prediction loss computed from clean stream logits. The two terms are weighted equally.

With this combined loss, the AR-equivalent signal fraction becomes:
\begin{equation}
    r_{\text{AR}}^{\text{(combined)}} = \frac{L + L/B}{2L} = \frac{1 + 1/B}{2}
    \label{eq:signal_ratio_sft}
\end{equation}
where the numerator counts $L$ pure AR terms from the clean stream plus $L/B$ AR-equivalent terms from the masked stream (one per block). For $B{=}4$, this is 62.5\%; for $B{=}16$, 53.1\%. In all cases the ratio stays above 50\%, effectively decoupling the AR signal from block size. The model simultaneously learns to predict from masked context \textit{and} maintains its original autoregressive competence, rather than replacing one with the other.

By default \ours{} includes the SFT loss. To isolate its effect, we also evaluate a variant trained without it (denoted ``\ours{} w/o SFT loss''), which uses $\mathcal{L}_{\text{mask}}$ alone.

\subsection{Inference: Left-to-Right Sliding Window}
\label{sec:method_inference}

The inference procedure completes the AR-consistent design by closing gap (4): tokens are always accepted strictly left-to-right, matching AR generation order. Combined with causal attention and right-shifted logits from training, the result is that \ours{} at inference is indistinguishable from AR generation when only one token is accepted per step, and smoothly extends to multi-token generation when the model is confident.

At each step, $B$ number of \mask{} tokens are appended after the current prefix and the model runs a single forward pass with pure causal attention to obtain logits for all $B$ positions. Starting from the leftmost masked position, tokens are accepted consecutively while $\max_v p(x_t {=} v) \geq \tau$, with at least one token always accepted. The $N$ accepted tokens join the prefix, and $N$ new \mask{} tokens are appended to keep the window at size $B$. This repeats until generation is complete. The guarantee that at least one token is always accepted ensures that the model degrades gracefully to standard AR decoding when no prediction is confident enough.

The threshold $\tau$ directly controls throughput: $\tau \to 1.0$ accepts at most one token per step (recovering exact AR behavior), while lower $\tau$ accepts more tokens per step at some quality cost. Crucially, $\tau$ can be adjusted \textit{on-the-fly per request} during serving, without retraining or loading a different model. Section~\ref{sec:analysis_tradeoff} characterizes the full tradeoff curve.

\section{Experiments}
\label{sec:experiments}

We organize experiments around three claims: \ours{} preserves the original AR model's quality (Section~\ref{sec:exp_preserve}), the SFT loss is necessary and sufficient for this preservation at scale (Section~\ref{sec:exp_signal}), multi-token generation provides a smooth and controllable speed--quality frontier (Section~\ref{sec:exp_tradeoff}), and these gains translate to wall-clock speedup in batch inference (Section~\ref{sec:exp_wallclock}).

\subsection{Setup}
\label{sec:exp_setup}

We evaluate at two scales: Qwen2.5-0.5B-Instruct and Qwen2.5-7B-Instruct~\citep{qwen2025qwen25technicalreport}, both trained on Dolci-Instruct-SFT~\citep{olmo2025olmo3} ($\sim$2M examples). We first train an AR SFT model for 5 epochs with standard next-token prediction, then continue with \ours{} training on the same data for another 5 epochs. At 0.5B we train with block sizes $B \in \{4, 8, 16\}$; at 7B we use $B{=}4$. Full hyperparameters are in Appendix~\ref{app:training_details}.

We compare against the AR SFT starting point and Block Diffusion~\citep{arriolablock} (0.5B only). Evaluation uses six benchmarks: IFEval (0-shot)~\citep{zhou2023instruction}, BBH (3-shot)~\citep{suzgun2022challengingbigbenchtaskschainofthought}, MMLU-Pro (0-shot)~\citep{wang2024mmluprorobustchallengingmultitask}, GPQA (0-shot)~\citep{rein2023gpqagraduatelevelgoogleproofqa}, GSM8K (0-shot)~\citep{cobbe2021training}, and HumanEval (0-shot)~\citep{chen2021evaluating}, all with greedy decoding and max 256 new tokens.

\subsection{\ours{} Preserves AR Quality in One-Token Mode}
\label{sec:exp_preserve}

\begin{table}[t]
\caption{One-token mode ($\tau{=}1.0$): \ours{} vs.\ AR SFT, compute-matched AR SFT (10 epochs), and Block Diffusion. All models generate one token per forward pass. \textbf{Bold}: best per column within each scale.}
\label{tab:main_results}
\centering
\resizebox{\linewidth}{!}{\begin{tabular}{lccccccc}
\toprule
\textbf{Model} & \textbf{IFEval} & \textbf{BBH} & \textbf{MMLU-Pro} & \textbf{GPQA} & \textbf{GSM8K} & \textbf{HumanEval} & \textbf{Avg} \\
\midrule
\multicolumn{8}{l}{\textit{Qwen2.5-0.5B-Instruct}} \\
\midrule
\textcolor{gray}{\textit{Base (no fine-tune)}} & \textcolor{gray}{27.8} & \textcolor{gray}{6.1} & \textcolor{gray}{2.4} & \textcolor{gray}{13.6} & \textcolor{gray}{27.4} & \textcolor{gray}{26.2} & \textcolor{gray}{17.2} \\
AR SFT (5 ep) & 48.4 & 26.3 & 11.9 & 17.9 & 32.0 & 35.4 & 28.7 \\
AR SFT (10 ep) & 47.8 & 26.3 & 9.3 & 14.1 & 28.3 & 32.3 & 26.4 \\
Block Diffusion~\cite{arriolablock} ($B{=}4$) & 47.1 & 7.5 & 2.0 & 17.9 & 30.6 & 31.7 & 22.8 \\
\ours-0.5B w/o SFT loss ($B{=}4$) & 48.5 & \textbf{27.4} & 12.3 & 19.0 & 29.5 & 33.5 & 28.4 \\
\ours-0.5B ($B{=}4$) & \textbf{51.3} & 26.6 & \textbf{12.4} & \textbf{19.4} & \textbf{32.8} & \textbf{40.2} & \textbf{30.4} \\
\midrule
\multicolumn{8}{l}{\textit{Qwen2.5-7B-Instruct}} \\
\midrule
\textcolor{gray}{\textit{Base (no fine-tune)}} & \textcolor{gray}{63.4} & \textcolor{gray}{24.0} & \textcolor{gray}{10.5} & \textcolor{gray}{11.6} & \textcolor{gray}{57.3} & \textcolor{gray}{82.9} & \textcolor{gray}{41.6} \\
AR SFT & 67.0 & 54.0 & 43.9 & \textbf{27.5} & 68.7 & 78.7 & 56.6 \\
\ours-7B ($B{=}4$) & \textbf{68.2} & \textbf{54.6} & \textbf{44.4} & 26.6 & \textbf{73.2} & \textbf{81.7} & \textbf{58.1} \\
\bottomrule
\end{tabular}}
\end{table}

The first claim is that \ours{} is a strict superset of AR: when generating one token per step, it should match or exceed the original model. Table~\ref{tab:main_results} confirms this at both scales. At 0.5B, \ours{} achieves 30.4 average versus AR SFT's 28.7, with a 4.8-point gain on HumanEval (40.2 vs 35.4). At 7B, \ours{} achieves 58.1 versus 56.6, with notable gains on GSM8K (+4.5) and HumanEval (+3.0). The multi-token capability comes at no cost to one-token quality; the additional masked-prediction training appears to act as a form of data augmentation that slightly \textit{improves} the AR mode.

To rule out that the gains simply arise from extra training compute, we include a compute-matched AR SFT baseline trained for 10 epochs (same total fine-tuning budget as \ours{}: 5 epochs AR + 5 epochs masked). Continuing AR SFT beyond 5 epochs actually \textit{hurts}: the average drops from 28.7 to 26.4, with MMLU-Pro falling from 11.9 to 9.3 and GSM8K from 32.0 to 28.3. The additional AR epochs overfit rather than improve, confirming that \ours{}'s gains come from the masked prediction objective, not from additional training alone.

Block Diffusion~\citep{arriolablock}, which uses bidirectional intra-block attention, tells the opposite story: it collapses on BBH (7.5) and MMLU-Pro (2.0), scores comparable to the untuned base model. This validates the gap analysis from Section~\ref{sec:gap_analysis}: not all block prediction formulations are compatible with AR pretraining. The ones that break causality, logits alignment, or generation order destroy the model's reasoning ability. \ours{} avoids all three.

\subsection{Why Larger Blocks Work: Validating the Signal Decay Hypothesis}
\label{sec:exp_signal}

\begin{table}[t]
\caption{Effect of SFT loss across block sizes (0.5B, $\tau{=}1.0$). Without the SFT loss, larger blocks rapidly erode quality; with it, performance is stable across $B$.}
\label{tab:signal_decay}
\centering
\resizebox{0.8\linewidth}{!}{\begin{tabular}{lccccccc}
\toprule
\textbf{Model} & \textbf{IFEval} & \textbf{BBH} & \textbf{MMLU-Pro} & \textbf{GPQA} & \textbf{GSM8K} & \textbf{HumanEval} & \textbf{Avg} \\
\midrule
\textcolor{gray}{\textit{AR SFT}} & \textcolor{gray}{48.4} & \textcolor{gray}{26.3} & \textcolor{gray}{11.9} & \textcolor{gray}{17.9} & \textcolor{gray}{32.0} & \textcolor{gray}{35.4} & \textcolor{gray}{28.7} \\
\midrule
\multicolumn{8}{l}{\textit{\ours{} w/o SFT loss}} \\
\midrule
$B{=}4$ & 48.5 & 27.4 & 12.3 & 19.0 & 29.5 & 33.5 & 28.4 \\
$B{=}8$ & 48.5 & 24.3 & 11.1 & 20.8 & 24.9 & 28.7 & 26.4 \\
$B{=}16$ & 42.6 & 21.7 & 10.6 & 16.3 & 21.0 & 20.7 & 22.2 \\
\midrule
\multicolumn{8}{l}{\textit{\ours{} with SFT loss}} \\
\midrule
$B{=}4$ & 51.3 & 26.6 & 12.4 & 19.4 & 32.8 & 40.2 & 30.4 \\
$B{=}8$ & 49.6 & 26.9 & 12.0 & 19.6 & 32.8 & 37.2 & 29.7 \\
$B{=}16$ & 50.7 & 27.0 & 12.1 & 17.9 & 33.8 & 36.6 & 29.7 \\
\bottomrule
\end{tabular}}
\end{table}

\begin{figure}[t]
    \centering
    \includegraphics[width=\textwidth]{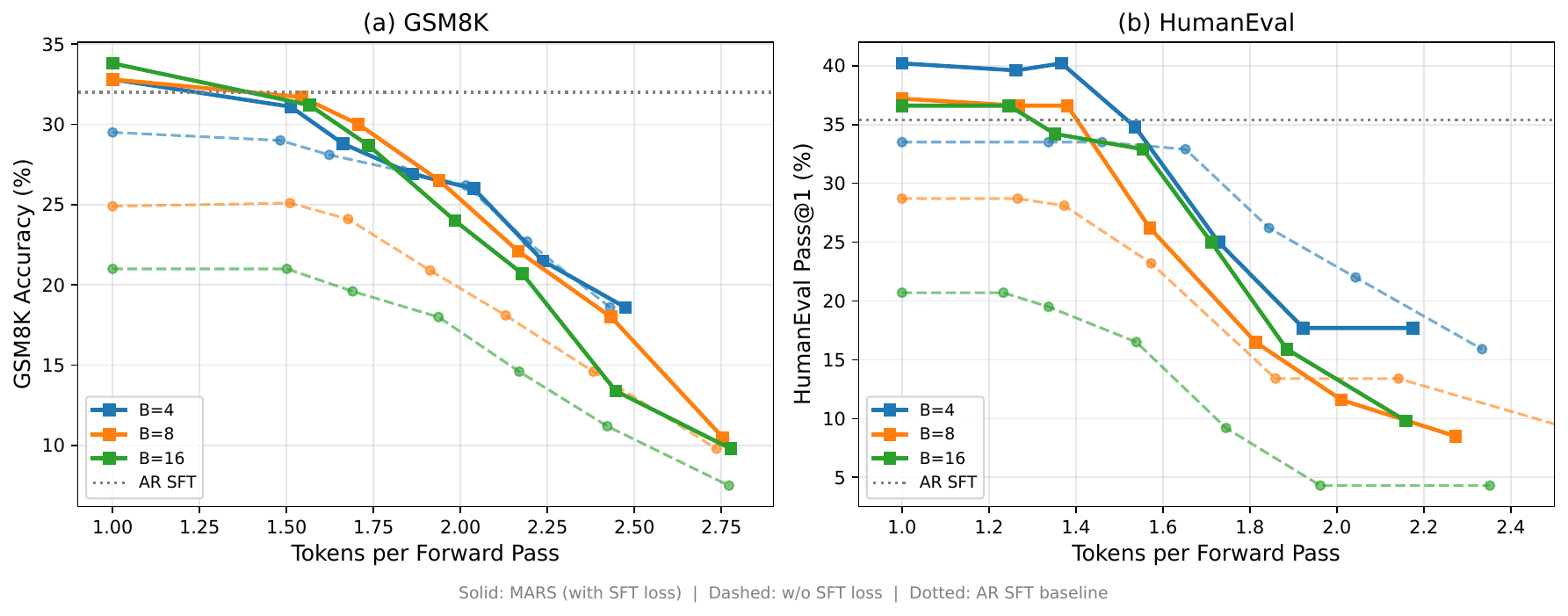}
    \caption{Speed--quality Pareto curves on GSM8K (left) and HumanEval (right). Solid lines: \ours{} (with SFT loss). Dashed lines: w/o SFT loss. Dotted: AR SFT baseline. With SFT loss, \ours{} dominates at every operating point on both tasks.}
    \label{fig:pareto}
\end{figure}

Section~\ref{sec:signal_decay} predicted that without the SFT loss, the AR training signal decays as $1/B$, causing larger blocks to degrade. Table~\ref{tab:signal_decay} confirms this directly.

Without the SFT loss, increasing block size from 4 to 16 drops the average from 28.4 to 22.2, a loss of 6.2 points. GSM8K falls from 29.5 to 21.0; HumanEval from 33.5 to 20.7. The degradation is systematic and concentrated in reasoning and coding tasks, exactly the capabilities most sensitive to the quality of the AR signal.

With the SFT loss, the same block size increase causes a drop of only 0.7 points (30.4 to 29.7). GSM8K actually \textit{improves} from 32.8 to 33.8, and HumanEval drops modestly from 40.2 to 36.6. The SFT loss stabilizes the AR signal ratio above 50\% regardless of $B$ (Eq.~\ref{eq:signal_ratio_sft}), and the empirical results match this prediction closely.

Figure~\ref{fig:pareto} shows the same pattern across the full threshold sweep: at every operating point (every tokens-per-forward rate), \ours{} with SFT loss achieves higher accuracy than without, and the gap widens at larger block sizes. The SFT loss does not just help at $\tau{=}1.0$; it lifts the entire Pareto frontier.

\subsection{A Smooth Speed--Quality Frontier via Confidence Thresholding}
\label{sec:exp_tradeoff}
\label{sec:analysis_tradeoff}

\begin{table}[t]
\caption{Multi-token mode. Each cell shows accuracy with the absolute change from $\tau{=}1.0$ to  $\tau{=}0.95$ in parentheses. Gray italic rows show tokens accepted per forward pass for each task. Full threshold sweep in Appendix Table~\ref{tab:threshold_sweep}.}
\label{tab:speedup}
\centering
\resizebox{\linewidth}{!}{\begin{tabular}{lccccccc}
\toprule
\textbf{Model} & \textbf{IFEval} & \textbf{BBH} & \textbf{MMLU-Pro} & \textbf{GPQA} & \textbf{GSM8K} & \textbf{HumanEval} & \textbf{Avg} \\
\midrule
\multicolumn{8}{l}{\textit{Qwen2.5-0.5B-Instruct}} \\
\midrule
\ours{} $B{=}4$ & 45.5\,\textcolor{gray}{\scriptsize($-$5.8)} & 26.6\,\textcolor{gray}{\scriptsize(0.0)} & 11.5\,\textcolor{gray}{\scriptsize($-$0.9)} & 19.9\,\textcolor{gray}{\scriptsize(+0.5)} & 31.1\,\textcolor{gray}{\scriptsize($-$1.7)} & 39.6\,\textcolor{gray}{\scriptsize($-$0.6)} & 29.0\,\textcolor{gray}{\scriptsize($-$1.4)} \\
\textcolor{gray}{\textit{tok/fwd}} & \textcolor{gray}{\textit{1.20}} & \textcolor{gray}{\textit{1.90}} & \textcolor{gray}{\textit{1.50}} & \textcolor{gray}{\textit{1.36}} & \textcolor{gray}{\textit{1.51}} & \textcolor{gray}{\textit{1.26}} & \textcolor{gray}{\textit{1.46}} \\
\ours{} $B{=}8$ & 44.0\,\textcolor{gray}{\scriptsize($-$5.6)} & 28.3\,\textcolor{gray}{\scriptsize(+1.4)} & 11.9\,\textcolor{gray}{\scriptsize($-$0.1)} & 19.0\,\textcolor{gray}{\scriptsize($-$0.6)} & 31.7\,\textcolor{gray}{\scriptsize($-$1.1)} & 36.6\,\textcolor{gray}{\scriptsize($-$0.6)} & 28.6\,\textcolor{gray}{\scriptsize($-$1.1)} \\
\textcolor{gray}{\textit{tok/fwd}} & \textcolor{gray}{\textit{1.23}} & \textcolor{gray}{\textit{1.96}} & \textcolor{gray}{\textit{1.49}} & \textcolor{gray}{\textit{1.46}} & \textcolor{gray}{\textit{1.54}} & \textcolor{gray}{\textit{1.27}} & \textcolor{gray}{\textit{1.49}} \\
\ours{} $B{=}16$ & 45.3\,\textcolor{gray}{\scriptsize($-$5.4)} & 25.9\,\textcolor{gray}{\scriptsize($-$1.1)} & 11.9\,\textcolor{gray}{\scriptsize($-$0.2)} & 17.9\,\textcolor{gray}{\scriptsize(0.0)} & 31.2\,\textcolor{gray}{\scriptsize($-$2.6)} & 36.6\,\textcolor{gray}{\scriptsize(0.0)} & 28.1\,\textcolor{gray}{\scriptsize($-$1.6)} \\
\textcolor{gray}{\textit{tok/fwd}} & \textcolor{gray}{\textit{1.21}} & \textcolor{gray}{\textit{1.86}} & \textcolor{gray}{\textit{1.51}} & \textcolor{gray}{\textit{1.43}} & \textcolor{gray}{\textit{1.57}} & \textcolor{gray}{\textit{1.25}} & \textcolor{gray}{\textit{1.47}} \\
\midrule
\multicolumn{8}{l}{\textit{Qwen2.5-7B-Instruct}} \\
\midrule
\ours{} $B{=}4$ & 63.0\,\textcolor{gray}{\scriptsize($-$5.2)} & 54.3\,\textcolor{gray}{\scriptsize($-$0.3)} & 44.2\,\textcolor{gray}{\scriptsize($-$0.2)} & 27.7\,\textcolor{gray}{\scriptsize(+1.1)} & 71.0\,\textcolor{gray}{\scriptsize($-$2.2)} & 80.5\,\textcolor{gray}{\scriptsize($-$1.2)} & 56.8\,\textcolor{gray}{\scriptsize($-$1.3)} \\
\textcolor{gray}{\textit{tok/fwd}} & \textcolor{gray}{\textit{1.15}} & \textcolor{gray}{\textit{2.60}} & \textcolor{gray}{\textit{1.47}} & \textcolor{gray}{\textit{1.47}} & \textcolor{gray}{\textit{1.79}} & \textcolor{gray}{\textit{1.60}} & \textcolor{gray}{\textit{1.68}} \\
\bottomrule
\end{tabular}}
\end{table}

Having established that \ours{} preserves AR quality, we now show that the same model provides a smooth, controllable tradeoff when multi-token generation is enabled. Table~\ref{tab:speedup} reports accuracy at $\tau{=}0.95$ alongside the change from one-token mode. Full threshold sweep results are in Table~\ref{tab:threshold_sweep} (Appendix).

The accuracy cost of multi-token generation is small and predictable. At 0.5B with $B{=}8$, the average drops by just 1.1 points (29.7$\to$28.6) while generating 1.49 tokens per forward pass. Most individual benchmarks lose less than 1 point. The largest drops are on IFEval ($\sim$5pp): IFEval evaluates strict adherence to formatting instructions (e.g., ``write exactly 3 paragraphs''), and multi-token acceptance tends to skip over format-critical tokens that the model would have produced more carefully in single-token mode. At 7B, the tradeoff is even more favorable: \ours-7B loses only 1.3 points on average (58.1$\to$56.8) while generating 1.68 tokens per forward, reaching 2.60 on BBH where the model produces high-confidence reasoning chains.

Crucially, the 7B model at $\tau{=}0.95$ (56.8) still exceeds the AR SFT baseline (56.6). The frontier is smooth with no cliff where quality suddenly collapses, giving the serving system fine-grained control. This is the ``opt-in'' property promised in Section~\ref{sec:introduction}: the same checkpoint serves quality-sensitive requests at $\tau{=}1.0$ and latency-sensitive requests at lower $\tau$, with no model swap and no retraining.

\subsection{Wall-Clock Speedup with Block-Level KV Cache}
\label{sec:exp_wallclock}
\label{sec:analysis_batch}

Tokens per forward pass measures algorithmic speedup, but wall-clock throughput is what matters in production. Standard AR decoding benefits from KV cache: each step processes only one new token. \ours{} processes $B$ masked tokens per step, so without caching, full-sequence recomputation makes it \textit{slower} than AR at batch sizes above 1.

\begin{figure}[t]
    \centering
    \includegraphics[width=\textwidth]{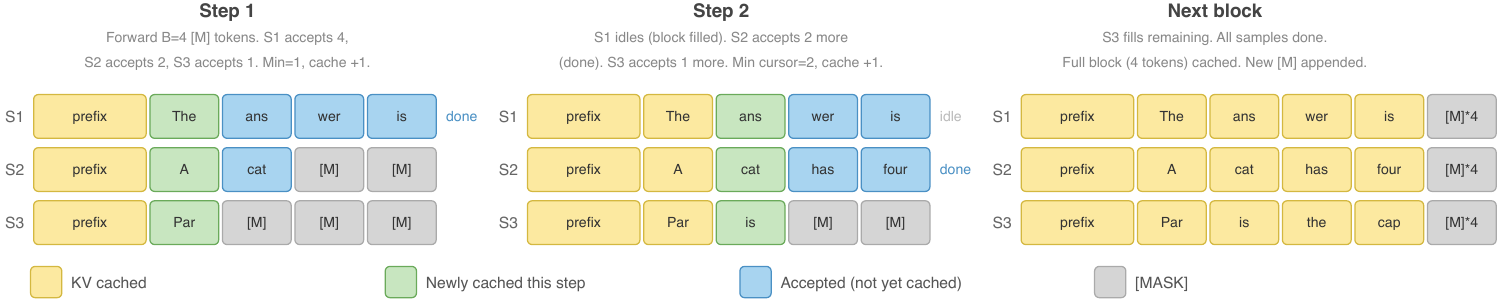}
    \caption{Block-level KV cache for batch inference ($B_{\text{cache}}{=}4$, batch size 3). Each step forwards $B$ \mask{} tokens against the cached prefix. The cache advances by the minimum number of tokens accepted across all samples: after Step 1 (S1 accepts 4, S2 accepts 2, S3 accepts 1), one token is cached (green). S1 idles while S2 and S3 continue. Once all samples fill the block, the entire block is cached (yellow) and new \mask{} tokens are appended for the next block.}
    \label{fig:block_cache}
\end{figure}

We implement a block-level KV cache strategy (Figure~\ref{fig:block_cache}): (1) compute the prefix KV cache once per block via a full forward pass, (2) iterate within the block using the cached prefix, where each inner step only forwards $B$ tokens, (3) once all samples in the batch have filled the block, extend the cache with the completed block and advance to the next. Faster samples idle at block boundaries until the slowest sample finishes. Table~\ref{tab:batch_inference} compares three configurations on GSM8K (256 questions) with Qwen2.5-7B at $\tau{=}0.95$.

\begin{table}[t]
\caption{Batch inference on GSM8K (256 questions) with Qwen2.5-7B at $\tau{=}0.95$. $B_{\text{cache}}$: KV cache synchronization granularity. For each batch size we report tokens/sec, total wall-clock time (seconds), and speedup relative to AR with KV cache ($\times$). \textbf{Bold}: best \ours{} configuration per batch size.}
\label{tab:batch_inference}
\centering
\small
\resizebox{\linewidth}{!}{\begin{tabular}{lr|rrr|rrr|rrr}
\toprule
 & & \multicolumn{3}{c|}{\textbf{Batch size=4}} & \multicolumn{3}{c|}{\textbf{Batch size=8}} & \multicolumn{3}{c}{\textbf{Batch size=16}} \\
\textbf{Method} & $B_{\text{cache}}$ & tok/s & time & $\times$ & tok/s & time & $\times$ & tok/s & time & $\times$ \\
\midrule
AR (KV cache) & -- & 143.4 & 276.2 & 1.00 & 236.7 & 169.1 & 1.00 & 434.3 & 91.8 & 1.00 \\
\midrule
\ours{} (no cache) & -- & 127.1 & 306.5 & 0.90 & 112.4 & 346.9 & 0.49 & 97.7 & 399.1 & 0.23 \\
\midrule
\ours{} + block cache & 4 & 157.2 & 248.8 & 1.11 & 236.4 & 165.4 & 1.02 & 399.3 & 97.2 & 0.94 \\
\ours{} + block cache & 8 & 196.0 & 199.4 & 1.39 & 289.9 & 134.4 & 1.26 & 484.1 & 80.1 & 1.15 \\
\ours{} + block cache & 16 & 228.5 & 170.5 & 1.62 & 338.0 & 115.1 & 1.47 & \textbf{566.0} & \textbf{68.7} & \textbf{1.34} \\
\ours{} + block cache & 32 & \textbf{241.9} & \textbf{161.2} & \textbf{1.71} & \textbf{368.4} & \textbf{105.6} & \textbf{1.60} & 544.9 & 71.5 & 1.28 \\
\ours{} + block cache & 64 & 239.5 & 162.0 & 1.70 & 342.6 & 113.8 & 1.49 & 383.8 & 101.5 & 0.90 \\
\ours{} + block cache & 128 & 194.2 & 200.7 & 1.38 & 217.9 & 178.8 & 0.95 & 223.9 & 174.0 & 0.53 \\
\ours{} + block cache & 256 & 122.7 & 317.6 & 0.87 & 128.7 & 302.7 & 0.56 & 124.1 & 313.9 & 0.29 \\
\bottomrule
\end{tabular}}
\end{table}

Block-level KV caching is essential for \ours{}: without it, throughput actually \textit{decreases} as batch size grows (127$\to$112$\to$98 tok/s) due to $O(T^2)$ full-sequence recomputation per step. With the cache, \ours{} outperforms AR at every batch size tested. At Batch size=4, the best configuration ($B_{\text{cache}}{=}32$) finishes in 161.2s versus AR's 276.2s, a \textbf{1.71$\times$} wall-clock speedup. At Batch size=8, \ours{} achieves \textbf{1.60$\times$} (105.6s vs 169.1s). At Batch size=16, \ours{} achieves \textbf{1.34$\times$} (68.7s vs 91.8s). The speedup is largest at smaller batch sizes where AR's per-token overhead is proportionally higher. The optimal cache granularity shifts with batch size ($B_{\text{cache}}{=}32$ at Batch size=4/8, $B_{\text{cache}}{=}16$ at Batch size=16), reflecting a tradeoff between amortizing prefix recomputation and synchronization overhead at block boundaries. Accuracy is preserved across all configurations (78--82\% GSM8K, within 2--3pp of AR).

\section{Conclusion}
\label{sec:conclusion}

We presented \ours{}, a lightweight fine-tuning method that gives instruction-tuned AR models the ability to generate multiple tokens per forward pass, with no architectural changes, no additional parameters, and a single checkpoint. The resulting model is a strict superset of the original: in one-token mode it matches or exceeds AR SFT at both 0.5B (+1.7 avg) and 7B (+1.5 avg), and in multi-token mode it achieves 1.5--1.7$\times$ throughput with minimal accuracy cost. With block-level KV caching, these gains translate to up to 1.71$\times$ wall-clock speedup over AR with KV cache on Qwen2.5-7B.

The method rests on two insights. First, block-masked prediction fails when it unnecessarily departs from AR behavior; closing the three eliminable gaps (attention pattern, logits alignment, generation order) is sufficient to recover baseline quality. Second, the SFT loss on the clean stream preserves the model's AR competence during masked-prediction training, preventing the AR signal from decaying as $1/B$ and stabilizing it above 50\% regardless of block size.

Due to computation constraints, we evaluate only $B{=}4$ for the 7B model. Our 0.5B experiments show that different block sizes yield similar Pareto frontiers on the speed--quality tradeoff (Table~\ref{tab:signal_decay}), with $B{=}4$ marginally ahead (less than 0.5pp in average accuracy). We expect this pattern to hold at larger scales, but verifying this remains future work. Other promising directions include: (1) cursor-based cache management that eliminates block-boundary synchronization, (2) adaptive block size selection based on input complexity, and (3) integration with speculative decoding for further acceleration.

\paragraph{Limitations.} \ours{} training concatenates a clean and noisy copy of each sequence, doubling the per-sample sequence length and therefore the training-time compute relative to standard SFT. This overhead is nonetheless lightweight compared to continual pretraining: training requires only 5 epochs of SFT data rather than large-scale pre-training corpora. The speed--quality tradeoff at aggressive thresholds ($\tau < 0.7$) shows substantial quality loss, suggesting room for improvement in the acceptance strategy. The block-level KV cache requires batch synchronization at block boundaries, which limits throughput gains at large batch sizes.

\bibliographystyle{plainnat}
\bibliography{references}

@article{chen2023accelerating,
  title={Accelerating large language model decoding with speculative sampling},
  author={Chen, Charlie and Borgeaud, Sebastian and Irving, Geoffrey and Lespiau, Jean-Baptiste and Sifre, Laurent and Jumper, John},
  journal={arXiv preprint arXiv:2302.01318},
  year={2023}
}

@inproceedings{leviathan2023fast,
  title={Fast inference from transformers via speculative decoding},
  author={Leviathan, Yaniv and Kalman, Matan and Matias, Yossi},
  booktitle={International Conference on Machine Learning},
  pages={19274--19286},
  year={2023},
  organization={PMLR}
}

@inproceedings{cai2024medusa,
  title={Medusa: Simple LLM Inference Acceleration Framework with Multiple Decoding Heads},
  author={Cai, Tianle and Li, Yuhong and Geng, Zhengyang and Peng, Hongwu and Lee, Jason D and Chen, Deming and Dao, Tri},
  booktitle={International Conference on Machine Learning},
  pages={5209--5235},
  year={2024},
  organization={PMLR}
}

@inproceedings{li2024eagle,
  title={EAGLE: Speculative Sampling Requires Rethinking Feature Uncertainty},
  author={Li, Yuhui and Wei, Fangyun and Zhang, Chao and Zhang, Hongyang},
  booktitle={International Conference on Machine Learning},
  pages={28935--28948},
  year={2024},
  organization={PMLR}
}

@inproceedings{arriolablock,
  title={Block Diffusion: Interpolating Between Autoregressive and Diffusion Language Models},
  author={Arriola, Marianne and Gokaslan, Aaron and Chiu, Justin T and Yang, Zhihan and Qi, Zhixuan and Han, Jiaqi and Sahoo, Subham Sekhar and Kuleshov, Volodymyr},
  booktitle={The Thirteenth International Conference on Learning Representations},
  year={2025}
}

@misc{zhou2026dllmsimplediffusionlanguage,
      title={dLLM: Simple Diffusion Language Modeling}, 
      author={Zhanhui Zhou and Lingjie Chen and Hanghang Tong and Dawn Song},
      year={2026},
      eprint={2602.22661},
      archivePrefix={arXiv},
      primaryClass={cs.CL},
      url={https://arxiv.org/abs/2602.22661}, 
}

@inproceedings{gongscaling,
  title={Scaling Diffusion Language Models via Adaptation from Autoregressive Models},
  author={Gong, Shansan and Agarwal, Shivam and Zhang, Yizhe and Ye, Jiacheng and Zheng, Lin and Li, Mukai and An, Chenxin and Zhao, Peilin and Bi, Wei and Han, Jiawei and others},
  booktitle={The Thirteenth International Conference on Learning Representations},
  year={2025}
}

@article{fu2025efficient,
  title={Efficient-dlm: From autoregressive to diffusion language models, and beyond in speed},
  author={Fu, Yonggan and Whalen, Lexington and Ye, Zhifan and Dong, Xin and Diao, Shizhe and Liu, Jingyu and Wu, Chengyue and Zhang, Hao and Xie, Enze and Han, Song and others},
  journal={arXiv preprint arXiv:2512.14067},
  year={2025}
}

@inproceedings{gu2018non,
  title={Non-Autoregressive Neural Machine Translation},
  author={Gu, Jiatao and Bradbury, James and Xiong, Caiming and Li, Victor OK and Socher, Richard},
  booktitle={International Conference on Learning Representations},
  year={2018}
}

@inproceedings{ghazvininejad2019mask,
  title={Mask-predict: Parallel decoding of conditional masked language models},
  author={Ghazvininejad, Marjan and Levy, Omer and Liu, Yinhan and Zettlemoyer, Luke},
  booktitle={Proceedings of the 2019 conference on empirical methods in natural language processing and the 9th international joint conference on natural language processing (EMNLP-IJCNLP)},
  pages={6112--6121},
  year={2019}
}

@article{sahoo2024simple,
  title={Simple and effective masked diffusion language models},
  author={Sahoo, Subham S and Arriola, Marianne and Schiff, Yair and Gokaslan, Aaron and Marroquin, Edgar and Chiu, Justin T and Rush, Alexander and Kuleshov, Volodymyr},
  journal={Advances in Neural Information Processing Systems},
  volume={37},
  pages={130136--130184},
  year={2024}
}

@inproceedings{lou2024discrete,
  title={Discrete Diffusion Modeling by Estimating the Ratios of the Data Distribution},
  author={Lou, Aaron and Meng, Chenlin and Ermon, Stefano},
  booktitle={International Conference on Machine Learning},
  pages={32819--32848},
  year={2024},
  organization={PMLR}
}

@article{ye2025dream,
  title={Dream 7b: Diffusion large language models},
  author={Ye, Jiacheng and Xie, Zhihui and Zheng, Lin and Gao, Jiahui and Wu, Zirui and Jiang, Xin and Li, Zhenguo and Kong, Lingpeng},
  journal={arXiv preprint arXiv:2508.15487},
  year={2025}
}

@inproceedings{nielarge,
  title={Large Language Diffusion Models},
  author={Nie, Shen and Zhu, Fengqi and You, Zebin and Zhang, Xiaolu and Ou, Jingyang and Hu, Jun and ZHOU, JUN and Lin, Yankai and Wen, Ji-Rong and Li, Chongxuan},
  booktitle={The Thirty-ninth Annual Conference on Neural Information Processing Systems},
  year={2025}
}

@misc{qwen2025qwen25technicalreport,
      title={Qwen2.5 Technical Report}, 
      author={Qwen and : and An Yang and Baosong Yang and Beichen Zhang and Binyuan Hui and Bo Zheng and Bowen Yu and Chengyuan Li and Dayiheng Liu and Fei Huang and Haoran Wei and Huan Lin and Jian Yang and Jianhong Tu and Jianwei Zhang and Jianxin Yang and Jiaxi Yang and Jingren Zhou and Junyang Lin and Kai Dang and Keming Lu and Keqin Bao and Kexin Yang and Le Yu and Mei Li and Mingfeng Xue and Pei Zhang and Qin Zhu and Rui Men and Runji Lin and Tianhao Li and Tianyi Tang and Tingyu Xia and Xingzhang Ren and Xuancheng Ren and Yang Fan and Yang Su and Yichang Zhang and Yu Wan and Yuqiong Liu and Zeyu Cui and Zhenru Zhang and Zihan Qiu},
      year={2025},
      eprint={2412.15115},
      archivePrefix={arXiv},
      primaryClass={cs.CL},
      url={https://arxiv.org/abs/2412.15115}, 
}

@misc{olmo2025olmo3,
      title={Olmo 3}, 
      author={Team Olmo and : and Allyson Ettinger and Amanda Bertsch and Bailey Kuehl and David Graham and David Heineman and Dirk Groeneveld and Faeze Brahman and Finbarr Timbers and Hamish Ivison and Jacob Morrison and Jake Poznanski and Kyle Lo and Luca Soldaini and Matt Jordan and Mayee Chen and Michael Noukhovitch and Nathan Lambert and Pete Walsh and Pradeep Dasigi and Robert Berry and Saumya Malik and Saurabh Shah and Scott Geng and Shane Arora and Shashank Gupta and Taira Anderson and Teng Xiao and Tyler Murray and Tyler Romero and Victoria Graf and Akari Asai and Akshita Bhagia and Alexander Wettig and Alisa Liu and Aman Rangapur and Chloe Anastasiades and Costa Huang and Dustin Schwenk and Harsh Trivedi and Ian Magnusson and Jaron Lochner and Jiacheng Liu and Lester James V. Miranda and Maarten Sap and Malia Morgan and Michael Schmitz and Michal Guerquin and Michael Wilson and Regan Huff and Ronan Le Bras and Rui Xin and Rulin Shao and Sam Skjonsberg and Shannon Zejiang Shen and Shuyue Stella Li and Tucker Wilde and Valentina Pyatkin and Will Merrill and Yapei Chang and Yuling Gu and Zhiyuan Zeng and Ashish Sabharwal and Luke Zettlemoyer and Pang Wei Koh and Ali Farhadi and Noah A. Smith and Hannaneh Hajishirzi},
      year={2025},
      eprint={2512.13961},
      archivePrefix={arXiv},
      primaryClass={cs.CL},
      url={https://arxiv.org/abs/2512.13961}, 
}

@article{zhou2023instruction,
  title={Instruction-following evaluation for large language models},
  author={Zhou, Jeffrey and Lu, Tianjian and Mishra, Swaroop and Brahma, Siddhartha and Basu, Sujoy and Luan, Yi and Zhou, Denny and Hou, Le},
  journal={arXiv preprint arXiv:2311.07911},
  year={2023}
}

@misc{suzgun2022challengingbigbenchtaskschainofthought,
  title={Challenging BIG-Bench Tasks and Whether Chain-of-Thought Can Solve Them},
  author={Mirac Suzgun and Nathan Scales and Nathanael Schärli and Sebastian Gehrmann and Yi Tay and Hyung Won Chung and Aakanksha Chowdhery and Quoc V. Le and Ed H. Chi and Denny Zhou and Jason Wei},
  year={2022},
  eprint={2210.09261},
  archivePrefix={arXiv},
  primaryClass={cs.CL},
  url={https://arxiv.org/abs/2210.09261},
}

@misc{rein2023gpqagraduatelevelgoogleproofqa,
  title={GPQA: A Graduate-Level Google-Proof Q\&A Benchmark},
  author={David Rein and Betty Li Hou and Asa Cooper Stickland and Jackson Petty and Richard Yuanzhe Pang and Julien Dirani and Julian Michael and Samuel R. Bowman},
  year={2023},
  eprint={2311.12022},
  archivePrefix={arXiv},
  primaryClass={cs.AI},
  url={https://arxiv.org/abs/2311.12022},
}

@misc{wang2024mmluprorobustchallengingmultitask,
  title={MMLU-Pro: A More Robust and Challenging Multi-Task Language Understanding Benchmark},
  author={Yubo Wang and Xueguang Ma and Ge Zhang and Yuansheng Ni and Abhranil Chandra and Shiguang Guo and Weiming Ren and Aaran Arulraj and Xuan He and Ziyan Jiang and Tianle Li and Max Ku and Kai Wang and Alex Zhuang and Rongqi Fan and Xiang Yue and Wenhu Chen},
  year={2024},
  eprint={2406.01574},
  archivePrefix={arXiv},
  primaryClass={cs.CL},
  url={https://arxiv.org/abs/2406.01574},
}

@article{cobbe2021training,
  title={Training verifiers to solve math word problems},
  author={Cobbe, Karl and Kosaraju, Vineet and Bavarian, Mohammad and Chen, Mark and Jun, Heewoo and Kaiser, Lukasz and Plappert, Matthias and Tworek, Jerry and Hilton, Jacob and Nakano, Reiichiro and others},
  journal={arXiv preprint arXiv:2110.14168},
  year={2021}
}

@article{chen2021evaluating,
  title={Evaluating large language models trained on code},
  author={Chen, Mark and Tworek, Jerry and Jun, Heewoo and Yuan, Qiming and Pinto, Henrique Ponde De Oliveira and Kaplan, Jared and Edwards, Harri and Burda, Yuri and Joseph, Nicholas and Brockman, Greg and others},
  journal={arXiv preprint arXiv:2107.03374},
  year={2021}
}

@inproceedings{tengaccelerating,
  title={Accelerating Auto-regressive Text-to-Image Generation with Training-free Speculative Jacobi Decoding},
  author={Teng, Yao and Shi, Han and Liu, Xian and Ning, Xuefei and Dai, Guohao and Wang, Yu and Li, Zhenguo and Liu, Xihui},
  booktitle={The Thirteenth International Conference on Learning Representations},
  year={2024}
}

@inproceedings{fu2024break,
  title={Break the Sequential Dependency of LLM Inference Using Lookahead Decoding},
  author={Fu, Yichao and Bailis, Peter and Stoica, Ion and Zhang, Hao},
  booktitle={International Conference on Machine Learning},
  pages={14060--14079},
  year={2024},
  organization={PMLR}
}

@article{ruan2026causal,
  title={Causal Autoregressive Diffusion Language Model},
  author={Ruan, Junhao and Li, Bei and Yin, Yongjing and Huang, Pengcheng and Chen, Xin and Wang, Jingang and Cai, Xunliang and Xiao, Tong and Zhu, JingBo},
  journal={arXiv preprint arXiv:2601.22031},
  year={2026}
}

@article{karami2026auto,
  title={Auto-Regressive Masked Diffusion Models},
  author={Karami, Mahdi and Ghodsi, Ali},
  journal={arXiv preprint arXiv:2601.16971},
  year={2026}
}

@article{du2026autoregressive,
  title={Autoregressive Models Rival Diffusion Models at ANY-ORDER Generation},
  author={Du, Tianqi and Fang, Lizhe and Yang, Weijie and Zhang, Chenheng and Wei, Zeming and Wang, Yifei and Wang, Yisen},
  journal={arXiv preprint arXiv:2601.13228},
  year={2026}
}

@misc{sahoo2026esotericlanguagemodelsbridging,
      title={Esoteric Language Models: Bridging Autoregressive and Masked Diffusion LLMs}, 
      author={Subham Sekhar Sahoo and Zhihan Yang and Yash Akhauri and Johnna Liu and Deepansha Singh and Zhoujun Cheng and Zhengzhong Liu and Eric Xing and John Thickstun and Arash Vahdat},
      year={2026},
      eprint={2506.01928},
      archivePrefix={arXiv},
      primaryClass={cs.CL},
      url={https://arxiv.org/abs/2506.01928}, 
}

@inproceedings{gloeckle2024better,
  title={Better \& Faster Large Language Models via Multi-token Prediction},
  author={Gloeckle, Fabian and Idrissi, Badr Youbi and Roziere, Baptiste and Lopez-Paz, David and Synnaeve, Gabriel},
  booktitle={International Conference on Machine Learning},
  pages={15706--15734},
  year={2024},
  organization={PMLR}
}

@article{liu2024deepseek,
  title={Deepseek-v3 technical report},
  author={Liu, Aixin and Feng, Bei and Xue, Bing and Wang, Bingxuan and Wu, Bochao and Lu, Chengda and Zhao, Chenggang and Deng, Chengqi and Zhang, Chenyu and Ruan, Chong and others},
  journal={arXiv preprint arXiv:2412.19437},
  year={2024}
}

@inproceedings{kou2024cllms,
  title={Cllms: Consistency large language models},
  author={Kou, Siqi and Hu, Lanxiang and He, Zhezhi and Deng, Zhijie and Zhang, Hao},
  booktitle={Forty-first International Conference on Machine Learning},
  year={2024}
}

@article{austin2021structured,
  title={Structured denoising diffusion models in discrete state-spaces},
  author={Austin, Jacob and Johnson, Daniel D and Ho, Jonathan and Tarlow, Daniel and Van Den Berg, Rianne},
  journal={Advances in neural information processing systems},
  volume={34},
  pages={17981--17993},
  year={2021}
}

@article{li2022diffusion,
  title={Diffusion-lm improves controllable text generation},
  author={Li, Xiang and Thickstun, John and Gulrajani, Ishaan and Liang, Percy S and Hashimoto, Tatsunori B},
  journal={Advances in neural information processing systems},
  volume={35},
  pages={4328--4343},
  year={2022}
}

@article{shi2024simplified,
  title={Simplified and generalized masked diffusion for discrete data},
  author={Shi, Jiaxin and Han, Kehang and Wang, Zhe and Doucet, Arnaud and Titsias, Michalis},
  journal={Advances in neural information processing systems},
  volume={37},
  pages={103131--103167},
  year={2024}
}

@article{gat2024discrete,
  title={Discrete flow matching},
  author={Gat, Itai and Remez, Tal and Shaul, Neta and Kreuk, Felix and Chen, Ricky TQ and Synnaeve, Gabriel and Adi, Yossi and Lipman, Yaron},
  journal={Advances in Neural Information Processing Systems},
  volume={37},
  pages={133345--133385},
  year={2024}
}

@article{wu2025fast,
  title={Fast-dllm: Training-free acceleration of diffusion llm by enabling kv cache and parallel decoding},
  author={Wu, Chengyue and Zhang, Hao and Xue, Shuchen and Liu, Zhijian and Diao, Shizhe and Zhu, Ligeng and Luo, Ping and Han, Song and Xie, Enze},
  journal={arXiv preprint arXiv:2505.22618},
  year={2025}
}

@article{gao2025self,
  title={Self speculative decoding for diffusion large language models},
  author={Gao, Yifeng and Ji, Ziang and Wang, Yuxuan and Qi, Biqing and Xu, Hanlin and Zhang, Linfeng},
  journal={arXiv preprint arXiv:2510.04147},
  year={2025}
}

\appendix

\section{Training Details}
\label{app:training_details}

\begin{table}[h]
\caption{Training hyperparameters. Both stages (AR SFT $\to$ \ours{}) use identical settings per model size. Effective batch sizes are matched across scales.}
\label{tab:hyperparameters}
\centering
\begin{tabular}{lll}
\toprule
\textbf{Hyperparameter} & \textbf{0.5B} & \textbf{7B} \\
\midrule
Base model & Qwen2.5-0.5B-Instruct & Qwen2.5-7B-Instruct \\
Training data & \multicolumn{2}{c}{Dolci-Instruct-SFT ($\sim$2M examples)} \\
Optimizer & \multicolumn{2}{c}{AdamW} \\
Learning rate & \multicolumn{2}{c}{$5 \times 10^{-6}$} \\
LR schedule & \multicolumn{2}{c}{Cosine decay} \\
Warmup ratio & \multicolumn{2}{c}{0.03} \\
Epochs & \multicolumn{2}{c}{5 (per stage)} \\
Max sequence length & \multicolumn{2}{c}{512} \\
Per-device batch size & 48 & 24 (AR) / 12 (\ours{}) \\
Gradient accumulation & 1 & 2 (AR) / 4 (\ours{}) \\
Effective batch size & 384 & 384 \\
Precision & \multicolumn{2}{c}{bfloat16} \\
Hardware & \multicolumn{2}{c}{8$\times$ NVIDIA H200} \\
Block sizes tested & 4, 8, 16 & 4 \\
SFT loss & \multicolumn{2}{c}{Included (\ours{}) / Excluded (\ours{} w/o SFT loss)} \\
\midrule
\multicolumn{3}{l}{\textit{Evaluation}} \\
\midrule
Generation length & \multicolumn{2}{c}{256 tokens} \\
Decoding & \multicolumn{2}{c}{Greedy (temperature = 0)} \\
Steps (\ours{}) & \multicolumn{2}{c}{256 (one token per step at $\tau{=}1.0$)} \\
\bottomrule
\end{tabular}
\end{table}

\paragraph{Training cost.} \ours{} training concatenates a clean and noisy copy of each sequence, doubling the effective sequence length. This incurs additional training cost: for the 0.5B model, AR SFT takes 15 H200-hours while \ours{} takes 33 H200-hours (2.2$\times$); for the 7B model, 100 vs 202 H200-hours (2.0$\times$). Peak GPU memory usage is approximately 1.5$\times$ that of AR SFT. This overhead is modest compared to continual pretraining: both stages use only 5 epochs of SFT data.

\section{Threshold Sweep Details}
\label{app:threshold_sweep}

Table~\ref{tab:threshold_sweep} reports the full speed--quality tradeoff for \ours{} and \ours{} w/o SFT loss across block sizes $B \in \{4, 8, 16\}$ on GSM8K and HumanEval, sweeping the acceptance threshold $\tau$ from 1.0 (one token per step, equivalent to AR) down to 0.5. At $\tau{=}0.95$, \ours{} with $B{=}4$ achieves 1.51 tokens per forward on GSM8K with only 1.7pp accuracy loss relative to $\tau{=}1.0$. Lowering $\tau$ further increases throughput but degrades quality, particularly for larger block sizes. The SFT loss variant consistently dominates the w/o SFT loss variant across all operating points, confirming that the SFT loss stabilizes performance across the entire Pareto frontier, not just at $\tau{=}1.0$.

\begin{table}[h]
\caption{Speed--quality tradeoff on GSM8K (0-shot) and HumanEval (0-shot). Tok/Fwd: average tokens accepted per forward pass. $\tau{=}1.0$: one token per step (from Table~\ref{tab:main_results}). Lower $\tau$ accepts more tokens but may reduce accuracy.}
\label{tab:threshold_sweep}
\centering
\small
\resizebox{\linewidth}{!}{\begin{tabular}{l|cc|cc|cc|cc|cc|cc}
\toprule
 & \multicolumn{6}{c|}{\textbf{GSM8K}} & \multicolumn{6}{c}{\textbf{HumanEval}} \\
 & \multicolumn{2}{c|}{$B{=}4$} & \multicolumn{2}{c|}{$B{=}8$} & \multicolumn{2}{c|}{$B{=}16$} & \multicolumn{2}{c|}{$B{=}4$} & \multicolumn{2}{c|}{$B{=}8$} & \multicolumn{2}{c}{$B{=}16$} \\
$\tau$ & Acc & T/F & Acc & T/F & Acc & T/F & P@1 & T/F & P@1 & T/F & P@1 & T/F \\
\midrule
\multicolumn{13}{l}{\textit{\ours{}}} \\
\midrule
\textcolor{gray}{\textit{AR SFT}} & \textcolor{gray}{32.0} & \textcolor{gray}{1.00} & \textcolor{gray}{32.0} & \textcolor{gray}{1.00} & \textcolor{gray}{32.0} & \textcolor{gray}{1.00} & \textcolor{gray}{35.4} & \textcolor{gray}{1.00} & \textcolor{gray}{35.4} & \textcolor{gray}{1.00} & \textcolor{gray}{35.4} & \textcolor{gray}{1.00} \\
1.0  & 32.8 & 1.00 & 32.8 & 1.00 & 33.8 & 1.00 & 40.2 & 1.00 & 37.2 & 1.00 & 36.6 & 1.00 \\
0.95 & 31.1 & 1.51 & 31.7 & 1.54 & 31.2 & 1.57 & 39.6 & 1.26 & 36.6 & 1.27 & 36.6 & 1.25 \\
0.9  & 28.8 & 1.66 & 30.0 & 1.71 & 28.7 & 1.74 & 40.2 & 1.37 & 36.6 & 1.38 & 34.2 & 1.35 \\
0.8  & 26.9 & 1.86 & 26.5 & 1.94 & 24.0 & 1.98 & 34.8 & 1.54 & 26.2 & 1.57 & 32.9 & 1.55 \\
0.7  & 26.0 & 2.04 & 22.1 & 2.17 & 20.7 & 2.18 & 25.0 & 1.73 & 16.5 & 1.81 & 25.0 & 1.71 \\
0.6  & 21.5 & 2.24 & 18.0 & 2.43 & 13.4 & 2.45 & 17.7 & 1.92 & 11.6 & 2.01 & 15.9 & 1.89 \\
0.5  & 18.6 & 2.47 & 10.5 & 2.75 &  9.8 & 2.78 & 17.7 & 2.18 &  8.5 & 2.27 &  9.8 & 2.16 \\
\midrule
\multicolumn{13}{l}{\textit{\ours{} w/o SFT loss}} \\
\midrule
1.0  & 29.5 & 1.00 & 24.9 & 1.00 & 21.0 & 1.00 & 33.5 & 1.00 & 28.7 & 1.00 & 20.7 & 1.00 \\
0.95 & 29.0 & 1.48 & 25.1 & 1.51 & 21.0 & 1.50 & 33.5 & 1.34 & 28.7 & 1.27 & 20.7 & 1.23 \\
0.9  & 28.1 & 1.62 & 24.1 & 1.68 & 19.6 & 1.69 & 33.5 & 1.46 & 28.1 & 1.37 & 19.5 & 1.34 \\
0.8  & 27.1 & 1.83 & 20.9 & 1.91 & 18.0 & 1.94 & 32.9 & 1.65 & 23.2 & 1.57 & 16.5 & 1.54 \\
0.7  & 26.2 & 2.02 & 18.1 & 2.13 & 14.6 & 2.17 & 26.2 & 1.84 & 13.4 & 1.86 &  9.2 & 1.75 \\
0.6  & 22.7 & 2.19 & 14.6 & 2.38 & 11.2 & 2.42 & 22.0 & 2.04 & 13.4 & 2.14 &  4.3 & 1.96 \\
0.5  & 18.6 & 2.43 &  9.8 & 2.74 &  7.5 & 2.77 & 15.9 & 2.33 &  9.2 & 2.53 &  4.3 & 2.35 \\
\bottomrule
\end{tabular}}
\end{table}

\section{Jacobi Decoding Baseline}
\label{app:jacobi}

\begin{table}[h]
\caption{Jacobi decoding~\citep{tengaccelerating} on the AR SFT checkpoint (0.5B, training-free). Jacobi uses fixed-point iteration: all future positions are initialized with random tokens and iteratively updated via causal forward passes until convergence. Tok/fwd: average tokens accepted per forward pass.}
\label{tab:jacobi}
\centering
\small
\resizebox{0.95\linewidth}{!}{
\begin{tabular}{lccccccc|c}
\toprule
\textbf{Method} & \textbf{IFEval} & \textbf{BBH} & \textbf{MMLU-Pro} & \textbf{GPQA} & \textbf{GSM8K} & \textbf{HumanEval} & \textbf{Avg} & \textbf{Tok/fwd} \\
\midrule
AR SFT & 48.4 & 26.3 & 11.9 & 17.9 & 32.0 & 35.4 & 28.7 & 1.00 \\
Jacobi (on AR SFT) & 41.0 & 19.2 & 11.7 & 17.6 & 36.5 & 42.1 & 28.0 & 1.07 \\
\midrule
\ours{} ($\tau{=}0.95$, $B{=}4$) & 45.5 & 26.6 & 11.5 & 19.9 & 31.1 & 39.6 & 29.0 & 1.46 \\
\bottomrule
\end{tabular}}
\end{table}

Table~\ref{tab:jacobi} compares Jacobi decoding with \ours{} on the same AR SFT checkpoint. Jacobi achieves only 1.07$\times$ average speedup (tokens per forward), compared to 1.46$\times$ for \ours{}. The limited speedup is expected: in Jacobi, each position sees random tokens as context for earlier positions in the generation region. Since the AR model was never trained to predict from incorrect prefixes, its predictions rarely converge in fewer iterations than sequential generation. \ours{} addresses this directly by training the model to predict from \mask{} placeholders, yielding substantially higher acceptance rates.

Jacobi does have one structural advantage: because it initializes all $N$ output positions at once, the model knows the exact generation length from the start, preventing it from generating beyond the intended boundary. This likely explains why Jacobi scores higher than AR SFT on GSM8K (36.5 vs 32.0) and HumanEval (42.1 vs 35.4), where output length control matters for correctness. On format-sensitive and reasoning tasks where this advantage does not apply, Jacobi drops significantly: IFEval ($-$7.4) and BBH ($-$7.1).

\section{Acceptance Metric Sensitivity}
\label{app:acceptance_metric}

The sliding-window inference by default accepts tokens left-to-right while a confidence score exceeds a threshold $\tau$. In the main experiments we use the probability of the top token, $\max_v p(v \mid \cdot)$, as the confidence score. Here we evaluate two alternatives on GSM8K with \ours{} ($B{=}4$):

\begin{itemize}[nosep,leftmargin=*]
    \item \textbf{Entropy}: $H(p) = -\sum_v p_v \log p_v$. Accept while $H \leq \tau$. Lower entropy indicates higher confidence.
    \item \textbf{Top-2 margin}: $p_{\text{top}_1} - p_{\text{top}_2}$. Accept while margin $\geq \tau$. A large gap between the best and second-best token indicates high confidence.
\end{itemize}

\begin{figure}[h]
\centering
\includegraphics[width=0.65\linewidth]{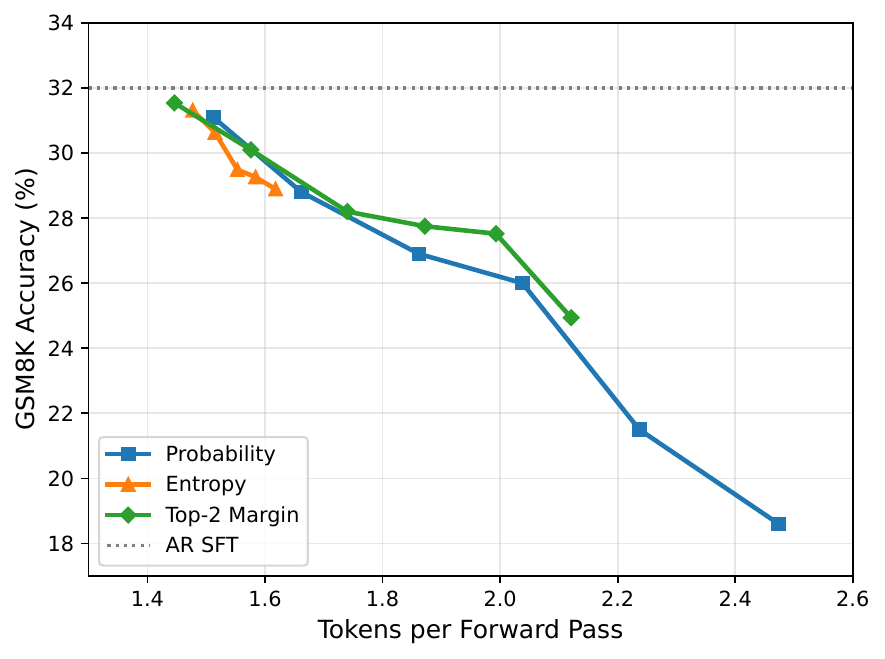}
\caption{Speed--quality trade-off under three acceptance metrics (MARS $B{=}4$, GSM8K). All three metrics trace similar Pareto frontiers, indicating that the speed--quality trade-off is robust to the choice of acceptance criterion. Entropy and top-2 margin degrade slightly more gracefully than raw probability at comparable tokens per forward pass.}
\label{fig:acceptance_metric}
\end{figure}

Figure~\ref{fig:acceptance_metric} shows the speed--quality frontier for each metric. All three trace similar Pareto curves, confirming that the sliding-window acceptance mechanism is robust to the specific confidence measure. Entropy and top-2 margin show marginally smoother degradation at comparable speedups, but the differences are small. We use probability in the main paper for its simplicity.

\end{document}